# IEEE Copyright Notice

# Design of an Autonomous Precision Pollination Robot

Nicholas Ohi[1], Kyle Lassak[1], Ryan Watson[1], Jared Strader[1], Yixin Du[2], Chizhao Yang[1], Gabrielle Hedrick[1], Jennifer Nguyen[1], Scott Harper[1], Dylan Reynolds[1], Cagri Kilic[1], Jacob Hikes[1], Sarah Mills[3], Conner Castle[1], Benjamin Buzzo[1], Nicole Waterland[3], Jason Gross[1], Yong-Lak Park[3], Xin Li[2], and Yu Gu[1]

*Abstract*— Precision robotic pollination systems can not only fill the gap of declining natural pollinators, but can also surpass them in efficiency and uniformity, helping to feed the fast-growing human population on Earth. This paper presents the design and ongoing development of an autonomous robot named "BrambleBee", which aims at pollinating bramble plants in a greenhouse environment. Partially inspired by the ecology and behavior of bees, BrambleBee employs state-of-the-art localization and mapping, visual perception, path planning, motion control, and manipulation techniques to create an efficient and robust autonomous pollination system.

## I. INTRODUCTION

An urgent issue faced by the agricultural sector today is the decline of natural pollinators, particularly honey bees, which threatens crop production. Many farmers cannot rely solely on natural pollinators in their local environments to effectively pollinate crops. Farmers often rent bees and have them shipped in from other locations for providing pollination services. In the United States alone, approximately $24 billion per year worth of crops depends on natural pollinators [1] and the declining bee population is increasing the cost to farmers who must rent them. Therefore, in parallel to addressing the cause of natural pollinator population decline (i.e. colony collapse disorder), there is a need to develop alternative pollination techniques to keep up with the increasing demands of the growing human population. One of these potential techniques is robotic precision pollination. Robotic pollinators can benefit the farmers by having more predictable availability than the insects, along with providing other functions such as flower thinning and crop data gathering.

Using robots to aid agriculture has been an active area of research for decades. Some key applications are fruit and vegetable picking [2-6], identifying and removing weeds [7-10], and mapping, phenotyping, and data collection of large fields of plants [11-14]. These technologies show how robots can contribute to the idea of "precision agriculture," where information about crops and their environment is used to make decisions about how to grow and maintain crops at an individual level and in a sustainable way [15-17]. In addition to information gathering, robots can greatly benefit precision agriculture by working to apply individualized treatments and maintenance. Humans may be able to do this effectively by hand in small-scale productions, but for large farms, robots could one day become much more efficient and cost-effective.

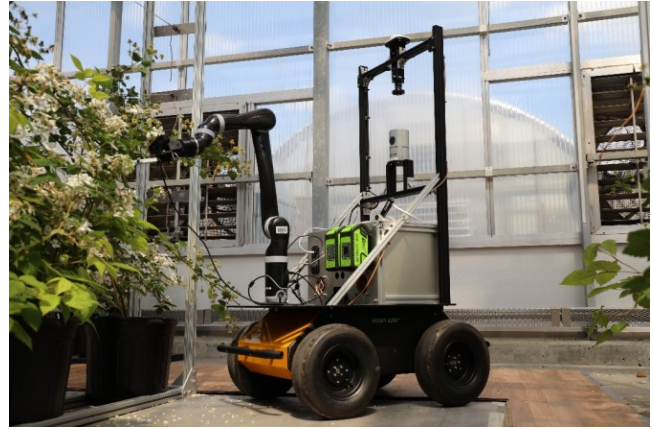

Figure 1. A precision pollination robot, BrambleBee, in the West Virginia University greenhouse with rows of bramble plants

The idea of using robots as pollinators, due to declining bee populations, has begun to gain popularity in recent years [18]. The precise and meticulous task of pollinating large numbers of flowers is very well suited to robots. The term "robotic pollination" likely conjures thoughts of small, bee-like robots flying around plants and going from flower to flower. Indeed, researchers have been investigating the design of small, insect-like, flying robots [19, 20] along with control policies for swarms of small flyers with a long-term aspiration of crop pollination [21]. Remotely controlled demonstrations of quadcopters that may be able to pollinate large flowers have also been performed [22]. Even though these kinds of demonstrations are at an early stage, they show the potential viability of robotic pollination. Flying robot based approaches, however, all face major challenges such as autonomy, duration, safety, and wind-disturbances to the flowers [23].

Examples of ground based robotic pollination systems include a fixed robotic crane used for vanilla pollination [24], a design for a mobile robotic arm for pollinating tall trees [25], and a mobile robot platform that pollinates tomato plants [26]. Beyond a handful of examples and conceptual designs, research into ground based robotic pollination systems is very limited and no significant autonomy has been demonstrated. Our research was intended to fill this gap. In this paper, an overview of a robotic system under development by researchers at West Virginia University (WVU) to perform fully autonomous precision pollination of bramble plants (i.e.,

N. Ohi; corresponding author; email: nohi@mix.wvu.edu
Y. Gu; email: yu.gu@mail.wvu.edu
1: Department of Mechanical and Aerospace Engineering, Benjamin M. Statler College of Engineering and Mineral Resources, West Virginia University (WVU), Morgantown, WV, 26506, USA.
2: Lane Department of Computer Science and Electrical Engineering, Benjamin M. Statler College of Engineering and Mineral Resources, WVU, Morgantown, WV, 26506, USA.
3: Division of Plant and Soil Sciences, Davis College of Agriculture, Natural Resources, and Design, WVU, Morgantown, WV, 26506, USA.

blackberries and raspberries) in a greenhouse is provided. The design of this robot, named "BrambleBee", shown in Figure 1, is described in the remaining sections as follows. Section II describes the similarities and differences between bees' and BrambleBee's pollination strategy; Section III provides an overview of BrambleBee's concept of operations; Section IV presents the design and instrumentation of BrambleBee; Section V describes the perception, localization, and mapping techniques used for robotic pollination; Section VI presents the robot autonomy approaches; Section VII describes the robotic manipulation system used to pollinate flowers; and Section VIII wraps up the paper with conclusions and plans for future work of this ongoing project.

## II. BEES AND BRAMBLEBEE

Bees forage for flowers primarily for the purpose of gathering food for themselves and their offspring [27], which provides pollination service as a byproduct. Many pollinating insects, including bees, tend to habituate and revisit known flower locations in an effort to minimize uncertainty in finding food [28]. This behavior is beneficial to the insects in terms of finding food, but could be detrimental in terms of pollination uniformity because not all flowers may get visited. Robotic pollinators, like BrambleBee, can be focused on pollination effectiveness and uniformity, rather than food gathering. Like bees, BrambleBee first finds flowers and then keeps track of their locations to plan its foraging path. BrambleBee's design for interacting with flowers is partially inspired by those of natural pollinators, such as mason bees (*Osmia* spp.), which collect pollen for their offspring. When mason bees land on flowers, their movement causes pollen to dislodge and the bees actively collect and attach the pollen to their scopa, which are bundles of fine hairs on their abdomens to hold collected pollen [29], as shown in Figure 2. As bees move across the flowers during foraging, some of the pollen sticks to the pistils (i.e., female reproductive organ) of flowers, resulting in pollination. Similarly, BrambleBee maneuvers its pollination mechanism, attached to the end of its robotic arm, to the flowers and uses precise motions to distribute pollen to the flowers' pistils, while avoiding damage to the flowers.

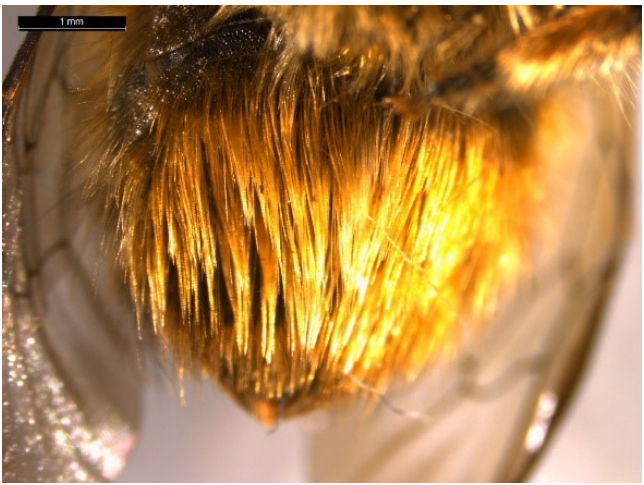

Figure 2. Microscopic image of a mason bee's scopa.

## III. ROBOT CONCEPT OF OPERATIONS

As shown in Figure 1 earlier, BrambleBee is a wheeled ground vehicle carrying a robotic arm with a pollination end-effector. It is also instrumented with a suite of sensors, including 3D Lidar, multiple cameras, wheel encoders, and an Inertial Measurement Unit (IMU).

Bramble plants are arranged in multiple rows inside of a rectangular greenhouse room with adequate spacing so that the robot can drive in between the rows and examine the plants from both sides. Prior knowledge about the geometry of the room and the arrangement of the plant rows is assumed to be roughly known to BrambleBee at the beginning of a session. BrambleBee then updates the map and determines the 3D locations of flower clusters as it moves in the environment. The map also contains a record of flower clusters' pollination history, estimates of pollination-readiness (i.e., how soon flower clusters will be ready to be pollinated) for newly identified clusters, and how long clusters will remain pollination-ready. All of this information is used to influence BrambleBee's planning and decision-making algorithms that determine where to visit to maximize its pollination efficiency over time.

At the beginning of each pollination session, BrambleBee drives around the room to inspect the plants and update its map. Detailed geometric mapping of the environment is performed using a Simultaneous Localization and Mapping (SLAM) algorithm. Semantic labeling of flower clusters is also performed using a combination of Lidar data and images from the onboard cameras. The readiness to be pollinated and remaining pollination viability time of each flower cluster is estimated from the image data which is then used by the planning algorithms to decide where to park the robot to maximize the number of unpollinated flowers that are within the reach of the robotic arm. Once the robot is parked, the arm with an RGB-D camera attached to its end-effector performs a scan to build up a high-resolution 3D map of the workspace with individual flower poses and conditions estimated. The arm then positions the pollination end-effector in front of each flower that needs to be pollinated, and uses visual-servoing to approach the flower. Bramble flowers can be self-pollinated, meaning pollen from a flower can be used to pollinate the same flower itself. Therefore, BrambleBee does not need to gather pollen from one flower and then deliver it to another. Rather, it must distribute the pollen on a flower evenly to the majority of the flower's pistils, so that most of the drupelets on the fruit develop properly and produce a full, well-formed berry. Once all flowers within reach are pollinated, BrambleBee then moves on to the next set of locations to pollinate other flowers.

## IV. ROBOT HARDWARE

BrambleBee is built primarily on a ClearPath Robotics® Husky platform. Mounted to the top front edge of the robot base is a Kinova® JACO$^2$ robotic arm and a custom end-effector used for precise flower pollination. The end-effector consists of a depth camera, a servo, three linear actuators to actuate the pollination tip of the end-effector precisely to touch individual flowers, and a small endoscope camera in the tip of the end-effector. The design of the pollination end-effector is inspired by the motion of bees when they land on flowers and the structure of the abdomens of bees, mimicking the hair-like scopa on their legs and abdomens that collect pollen from the flower. After field observations, it was found that most bees tend to land near the outside of the flower, brushing up against the anthers (i.e., male reproductive organ), collecting pollen, and then move towards the inner portion of the flower, containing the pistils. To achieve a similar action while minimizing the risk of damage to the flower, the pollinating end-effector design illustrated in Figure 3 was developed. It consists of three linear actuators that push and pull a flexible member on which cotton-tipped brushes are inserted. In the middle of the flexible member is an opening where the lens of an endoscope camera is inserted to aid the robotic arm with approaching the flower. When the linear actuators are extended, they will cause the flexible member to bend concave in relation to the flower, brushing against the anthers, toward the pistils. When the actuators retract, the flexible member will be convex in relation to the flower, moving the brushes out of view of the camera. Each actuator can be controlled independently to create a variety of brushing motions on the flower, while keeping the rest of the arm stationary. Currently, three iterations of this prototype have been built and are being tested on bramble flowers.

BrambleBee is also equipped with a variety of sensors to enable accurate perception of the surrounding environment to support intelligent decisions about where to go to pollinate flowers. One key sensor is a Velodyne® HDL-32E 32-channel 3D Lidar, which is used for robot localization, mapping, and obstacle avoidance. A FLIR® 5MP Blackfly-S camera outfitted with a fisheye lens is located directly above the Lidar and is used for mapping and flower cluster detection. A Novatel® SPAN GNSS/INS system is also installed onboard, but is mainly used for providing raw inertial measurements.

The quality of GNSS measurements inside a greenhouse environment is poor, so BrambleBee does not rely on GNSS for navigation due to extreme multipath and signal occlusion, but it collects GNSS data as it operates in the greenhouse to support research into the use of GNSS in such harsh environments. Located on the end of the robotic arm, in an "*eye-in-hand*" configuration, is an Intel® RealSense™ D400 depth camera, which is used for workspace mapping and visual guiding of the end-effector motion, and an endoscope camera in the center of the end-effector, used for additional guidance of the end-effector motion during close proximity to the flower. Wheel odometry measurements and motor current draw are available from the Husky platform and the positions of each joint of the robotic arm are available through sensors built into the arm.

The main computer and power distribution hardware are contained in a custom electronics box on top of the main plate of the Husky drive base. The drive base is powered by the Husky provided 24 V, 480 W·hr sealed lead acid battery and the sensors and computers are powered by up to four 40 V, 146 W·hr lithium-ion batteries. Figure 4 shows an image of BrambleBee, with all the major components labeled.

BrambleBee's main computer consists of an x86-64 motherboard with an Intel® Core i7-7700 3.6 GHz CPU and an Nvidia® GTX 1050 Ti GPU, running Ubuntu 16.04, and software written in C++ using the Robot Operating System (ROS). Custom printed circuit boards, derived from designs used by a previous sample return robot, Cataglyphis [30, 31], are used for power distribution and sensor interfacing. The current configuration of the electronics box can be seen in Figure 5.

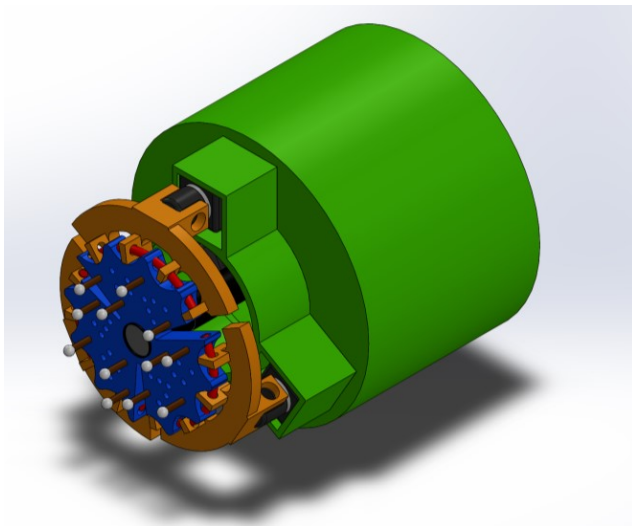

Figure 3. 3D CAD representation of the pollinating end-effector design. The blue component is 3D printed flexible thermoplastic polyurethane (TPU), the orange and green components are 3D printed rigid polylactic acid (PLA), and the grey and black components around the rim are parts of the three linear actuators and the lens of the endoscope camera is visible in the center of the flexible component. The brown and white components are cotton tipped brushes for pollen collection/delivery.

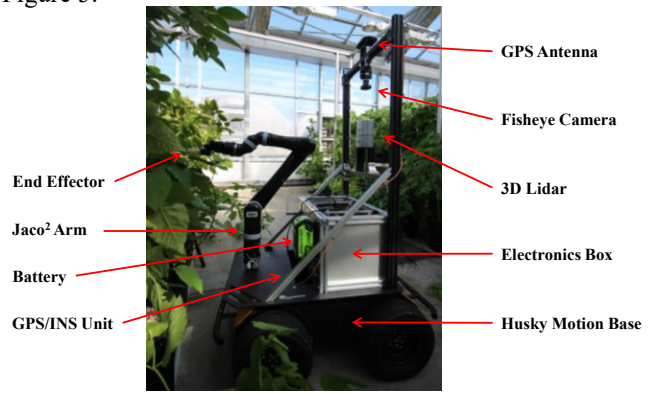

Figure 4. BrambleBee robot system with labeled components

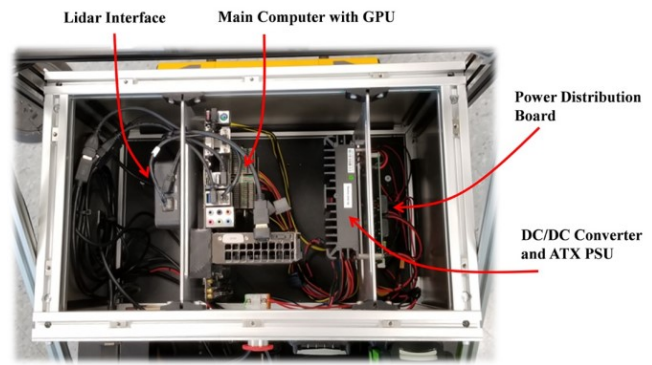

Figure 5. BrambleBee electronics box

## V. ROBOT PERCEPTION

### A. Localization and Mapping

Fast, accurate, and robust robot localization in a greenhouse environment is a key capability for BrambleBee. Localization is performed primarily with a real-time 3D SLAM algorithm. A factor graph based framework [32] is used to fuse the data from the sensor suite on-board BrambleBee to estimate the robot platform's states (i.e., position, orientation, and velocity).

Utilizing the factorization of the posterior distribution and the assumption that the system adheres to a Gaussian noise model, the state estimation problem simplifies to Non-Linear Least Squares (NLLS) optimization. Thus, any NLLS optimizer can be utilized, such as Levenberg-Marquardt [33]. This optimization framework is updated incrementally through the utilization of the Bayes' tree [34]. The Bayes' tree can be obtained from the factor graph through a two-step procedure, as described in [35]. The utilization of the Bayes' tree enables real-time optimization to be performed on-board BrambleBee. The Georgia Tech Smoothing and Mapping (GTSAM) library [36] is utilized to implement the incremental graph based optimization.

To improve the reliability of the loop-closure detection, each incoming point cloud scan from the 3D Lidar is matched with a prior 3D map of the greenhouse room using the Generalized Iterative Closest Point (Generalized-ICP) algorithm [37]. If the percentage of the matching points in the incoming scan is above a user-defined threshold, the loop-closure detection will be marked as successful. A transformation of the robot's current pose with respect to the original points of the prior map's coordinate frame (i.e., global frame) is then added to the factor graph as an observation link. The factor graph is then optimized based on the new observation link. A 2D occupancy grid map is also generated from the 3D SLAM map for path planning. An example SLAM solution is shown in Figure 6.

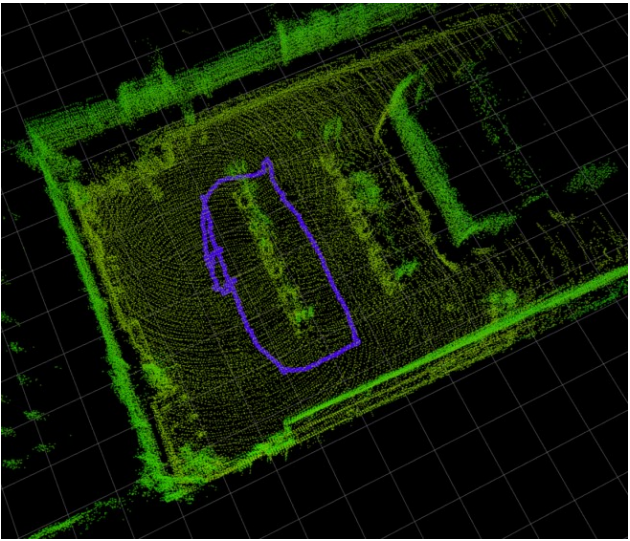

Figure 6. SLAM result showing the 3D map of the greenhouse generated by SLAM. The blue line shows the path of the robot as it performed an inspection path around one row of plants and then drove to two pollination locations.

Finally, to enable the robot to start navigating from anywhere in the greenhouse, a consistent global coordinate frame must be maintained to aid path planning. Therefore, the robot's initial offset, with respect to the origin of the global coordinate frame, located in one of the corners of the room, must be estimated. To obtain this initial offset, the 3D SLAM map is anchored to the prior map using the Generalized-ICP algorithm in a similar manner to the loop-closure detection described above. The initial offset is estimated every time the SLAM map is updated. Utilizing the most recent initial offset estimate, the states estimated in the BrambleBee local coordinate frame can be transformed into the global frame.

BrambleBee can estimate its states in real-time as it drives around the greenhouse with sufficient accuracy to prevent it from colliding with obstacles or getting lost. Incremental updates from wheel odometry and IMU data are used in between SLAM updates to provide high-rate pose estimates for real-time feedback control while driving. BrambleBee also estimates its initial pose in the global frame reliably from most starting poses. The next steps are to improve the reliability of this initial pose estimation, to continue improving the accuracy of the state estimation, and to improve the speed and accuracy of point cloud matching and verification for finding the initial pose estimate.

### B. Flower Detection

In addition to reliable robot pose estimation, BrambleBee must also be able to accurately identify and estimate the pose of flowers for pollination. This is accomplished using computer vision techniques with the downward-looking fisheye camera for the initial, long-range identification and then with the RGB-D camera on the robotic arm for precise, short-range positioning of the flowers.

The long-range algorithm can be separated into two parts. First, a rough segmentation is performed based on color to extract parts of the image that are most likely to be part of a flower. This process is completed using a naive Bayes' classifier [38] to assign pixels as belonging to or not belonging to flowers. In general, the naive Bayes' classifier is a family of conditional probability models based on applying Bayes' theorem with the assumption of conditional independence among features. Despite the oversimplified assumption of the naive Bayes' classifier, the initial rough segmentation is critical in reducing the overall processing time for each image. In this case, the pixel intensities are the features, and the classification rule is given by

$$\hat{c} = \arg\max_{c \in C} P(c)P(r|c)P(g|c)P(b|c) \quad (1)$$

where $C \in \{0,1\}$ is the set of labels (e.g. flower and non-flower) and $r$, $g$, and $b$ are the raw pixel intensities for each color channel (i.e. red, green, and blue).

To accelerate the segmentation process for real-time performance, a lookup table is pre-computed using all representable values for a pixel (e.g. 24 bits for most RGB images), and the lookup table is then accessed using the raw pixel values to efficiently compute the segmented image. The resulting segmentation produces a number of false positives due to shared colors between flowers and other objects. Thus, an approach based on transfer learning is adopted to distinguish between true and false positives from the initial segmentation. Google's Inception-v3 [39] was chosen for this task, because it performs well on the ImageNet benchmark dataset [40]. In this application, however, the objective is to distinguish true and false positives, so the original softmax layer, which has 1,000 classes, was modified to perform binary classification. The training took around 20 minutes using an Intel i7-4790k CPU and an NVIDIA Titan X GPU in Tensorflow. Training and testing statistics, as well as the recognition accuracy, are presented in Table I.

After the flowers are identified using the described approach, they are positioned in the map using unit vectors pointing from the camera to each flower. This process is described further in Section VI. The calibration procedure and camera model used for computing the unit vectors are provided in [41]. An example image segmented and the resulting patches classified are presented in Figure 7.

The short-range algorithm, using the RGB-D camera on the arm, consists of performing a reconstruction of the plant using real-time dense SLAM as presented in [42]. After the plant is reconstructed, the previously presented classification algorithm is used to identify the flowers in the series of images used for reconstruction. If a flower is identified, the position of the flower on the plant is obtained from the corresponding points in the reconstruction. The flower position estimate is then used to support manipulator control.

To develop computer vision algorithms, it is essential to collect a sufficient amount of image data for testing and validation. The data was collected at a local berry farm for various growth stages of the plants. Most of the data was collected with partially and fully bloomed flowers. The data was collected with multiple cameras for close-range and long-range images. For each single flower, the pose and growth stage of a flower is specified using a set of reference images. Over 500 images, each with many flowers, were used for training the segmentation and classification algorithms.

The next steps for computer vision research are to augment the classifier to accurately estimate flower poses to aid visual-servoing for the arm and to work on reliably differentiating individual flowers when many flowers are clustered together closely with large amounts of overlap. Another objective is to estimate the stage of development of detected flowers, which would also be recorded alongside the locations of detected flowers to be used for autonomous task planning that spans over multiple days of pollination. Also, the mapping of detected flowers will be improved by combining flower detection with the 3D SLAM to create a 3D semantic map of flower cluster locations, along with information about their estimated stages of development and history of whether they have been previously pollinated. This semantic map will be used in the future for planning where to pollinate more effectively, as described in the future work of Section VI, next.

TABLE I. TRAINING AND TESTING STATISTICS FOR FLOWER RECOGNITION, SHOWING THE NUMBER OF TRAINING AND TESTING PATCHES, AND THE NUMBER OF CORRECTLY CLASSIFIED PATCHES, ALONG WITH PRECISION AND RECALL.

|  | Training patches | Testing patches | Correctly Recognized patches | Precision | Recall |
|---|---|---|---|---|---|
| Flower | 13,395 | 2,102 | 1,892 | 78.63% | 90% |
| Non-flower | 15,066 | 2,124 | 1,609 | | |

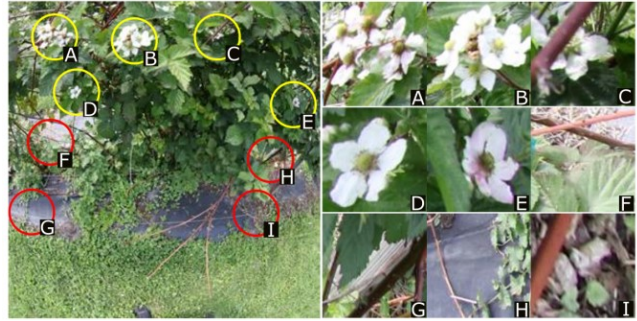

Figure 7. Example of the parts of the image extracted using the segmentation algorithm, where A-E were classified as flowers and F-I were classified as not flowers.

VI. AUTONOMY

Given information about the locations of flower clusters and the robot's location within the environment, BrambleBee must be able to make efficient plans to pollinate the plants in the greenhouse. First, to obtain up to date information about flower cluster locations and pollination readiness, BrambleBee must make its "inspection pass" of the greenhouse. The inspection path is generated by first discretizing the greenhouse environment into a graph Voronoi diagram and then solving for a path through the nodes in the graph. The path is found using visibility graph techniques and constrained so that the robot inspects all rows of plants while a balance between distance driven and safety from obstacles is maintained [43-46]. An example path to inspect three rows of plants is shown in Figure 8.

As BrambleBee drives the inspection path, nearby flower clusters are detected using the on-board fisheye camera, as described previously in Section V, Part B. The locations of the detected clusters are then recorded into a map of the plant rows by finding the intersections of the rays pointing in the directions of the detected clusters from the camera on the robot and the plant rows. Each row of plants is discretized into an equal number of "grid cells" on both sides of the rows. The number of flower clusters detected, that intersect a particular grid cell, is updated when the robot is within reliable visible range of that cell. Plant rows in the WVU greenhouse are approximately 3.44 m long and are divided into five equal length grid cells, each 0.688 m long.

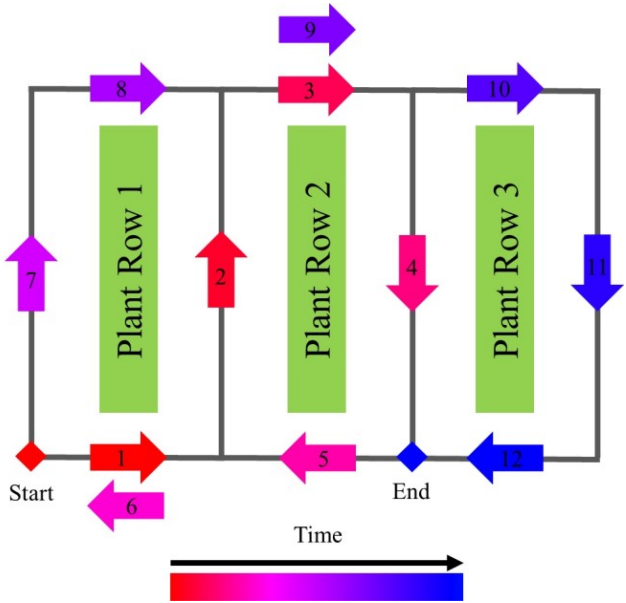

Figure 8. Graph Voronoi diagram path planning to inspect three discretized rows of plants. The color gradient, progressing from red to blue, shows the path of the robot over time. The robot starts at the red diamond and ends at the blue diamond, traversing segments to eventually cover all sides of all rows of plants.

After the inspection phase is completed and flower locations have been identified, BrambleBee then proceeds to decide where to go to pollinate flowers. Pollination locations are chosen by balancing the number of reachable flower clusters ready to be pollinated with minimizing distance driven in the robot drive base's configuration space. Pollination poses are selected from a set of poses associated with the grid cells in the map of plant rows. Each grid cell has a pre-computed "robot parking pose" which allows the arm to reach all spaces within that grid cell. The order to visit pollination locations is found using a greedy heuristic that chooses the next best grid cell to pollinate by selecting the one with the minimum cost, computed as

$$\arg\min_{k \in K} C_d \|x_{err}(k)\|_2 + C_f \frac{1}{N_f(k)} \quad (2)$$

where $\|x_{err}(k)\|_2$ is the Euclidean distance from the robot's current pose to the *k-th* grid cell's parking pose, $N_f(k)$ is the number of flower clusters contained in the *k-th* grid cell, and $C_d$ and $C_f$ are scaling factors for the distance and number of flowers terms, respectively.

Paths are then planned to efficiently reach these locations, while avoiding obstacles, using Dijkstra's algorithm for high level path planning and the Dynamic Window Approach for local obstacle avoidance [47]. Both algorithms use a dynamic occupancy grid map of the greenhouse created in real time from SLAM as described in Section V, Part A.

As the development of BrambleBee continues, the immediate next steps are to explore optimal solutions to the problem of choosing pollination parking locations, rather than using a greedy heuristic, and to incorporate models of the time-varying nature of flowers that become ready to be pollinated at different times. Also, the process of selecting pollination locations will be improved by reformulating it as a spatiotemporal planning problem, making use of both the locations of flower clusters in the 3D map and the clusters' varying stages of development, along with a model of how the flowers develop over time. Plans will be generated that span over multiple days, ensuring that flowers that are near the end of their pollination viability are pollinated sooner than flowers that will remain viable for longer. Finally, a longer-term goal is to unite the survey and the pollination phases into a combined task and motion planning algorithm where the robot autonomously decides when and where to search for flowers versus stopping to pollinate.

## VII. MANIPULATION

Once the robot arrives at a pollination location, BrambleBee must then use its robotic arm to precisely map individual flowers in the local workspace and then plan and execute an efficient sequence of actions to pollinate all reachable flowers. The workspace mapping is performed by moving the end-effector of the robotic arm in a survey pattern around the nearby plant while the poses of detected flowers are estimated and recorded into a local database using the depth camera on the end of the arm. Once all flowers in the local workspace have been identified, the sequence of flowers to pollinate is chosen by finding the sequence that minimizes the total distance traveled in the robot arm's joint configuration space. Currently, the shortest path is found using brute-force search; however, this is not feasible for large numbers of flowers. Therefore, planning heuristics will be applied to find a locally minimum shortest path, which will be the subject of future work. After the sequence of flowers to be pollinated has been determined, collision-free paths to reach observation positions directly in front of each flower are planned using the Open Motion Planning Library (OMPL) [48] and the Flexible Collision Library (FCL) [49]. The inverse kinematics and motion control of the arm are performed using the MoveIt! software package [50]. Once the end-effector arrives at one of the goal destinations, it is then parked in front of the flower, ready to perform the precise final approach and pollinate the flower. During the final approach maneuver, visual-servoing is performed to guide the tip of the end-effector precisely into the flower. The inverse kinematics for the final approach are solved using the TRAC-IK library [51].

Currently, testing of the manipulator path planning is being performed using easy to detect ArUco binary image markers [52, 53] in place of flowers. This is done to decouple the computer vision task of correctly identifying flowers and estimating their poses from the development of the arm planning task. Once reliable flower pose estimation has been achieved these two components will be combined and arm maneuvers will be made to pollinate bramble flowers. The arm as it "pollinates" an ArUco marker is shown in Figure 9.

Previously, an 18 cm "blind drive" in which the pollinator relied on the last best estimate of the marker pose, was required. This was because the depth camera could not see the whole marker as it got very close and made contact with the end-effector.

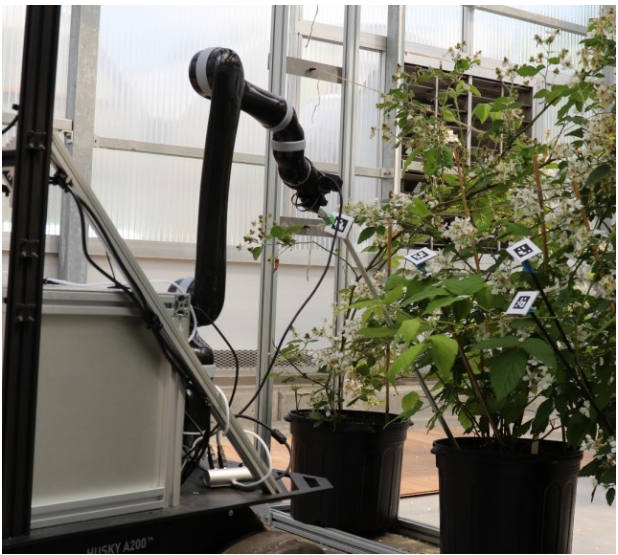

Figure 9. BrambleBee's arm, as it performs visual servoing to "pollinate" an ArUco binary image used for testing.

The updated end-effector design, presented in Section IV, mitigates this problem by employing an endoscope camera that is centrally located in the end-effector, which will allow for continuous tracking of the flower until contact. Additionally, planning constraints will be added so that the camera mounted on the end-effector can continuously track the next flower in the sequence during approach. This will increase the accuracy of the pose estimates of the flowers, and reduce the probability that a flower is lost during approach maneuvers.

## VIII. CONCLUSIONS AND FUTURE WORK

This paper presented an overview of BrambleBee, an autonomous precision robotic pollination system being developed at WVU to pollinate bramble flowers in a greenhouse environment. BrambleBee combines high accuracy mapping and localization techniques with visual flower identification and tracking, enabling efficient and robust planning and motion control to precisely and reliably pollinate bramble flowers. BrambleBee is the first of its kind in robotic pollination systems. It integrates several technologies to create an autonomous pollination system that can be used to help growers solve their pollination challenges in the near-term. In addition, the precise plant detection and manipulation capabilities gained during the development of BrambleBee can be used to enable other agricultural applications such as harvesting, pruning, and fruit picking.

A demonstration of the current capabilities of BrambleBee is shown in this video: https://youtu.be/66isrgth7-Q It shows that robotic pollination is now becoming a feasible concept, and given continuous development, will reach substantial capabilities. Future work to be completed has been discussed throughout the paper and is summarized here. This includes improving initial robot pose estimation using the prior map, reliably detecting the pose of individual flowers in dense flower clusters, optimizing planning algorithms for both the drive base and the arm, extending autonomy capabilities to make flexible pollination decisions, and realize the final sequence of pollination on real flowers.

ACKNOWLEDGMENT

This research was supported in part by USDA NIFA Project: 2017-67022-25926 and NSF GRFP Grant No. DGE-1102689.